\documentclass{article}

\usepackage[preprint,nonatbib]{neurips_2024}

\usepackage[utf8]{inputenc} 
\usepackage[T1]{fontenc}    
\usepackage{hyperref}       
\usepackage{url}            
\usepackage{booktabs}       
\usepackage{amsfonts}       
\usepackage{nicefrac}       
\usepackage{microtype}      
\usepackage{xcolor}         
\usepackage{graphicx}
\usepackage{amsmath}
\usepackage{amssymb}
\usepackage{multirow}
\usepackage{bbding}
\usepackage{booktabs}
\usepackage{bbm}
\usepackage{colortbl}
\usepackage{changepage} 
\usepackage{caption}
\usepackage{placeins}
\usepackage{titlesec}
\usepackage{wrapfig}
\usepackage{floatflt}
\usepackage{enumitem}

\titlespacing{\paragraph}{%
  0pt}{  
  -0.25em}{ 
  0.5em}   

\newcommand{\etal}{\textit{et al}.}

\title{Tuning-free Universally-Supervised  \\ Semantic Segmentation}

\author{%
  Xiaobo Yang\\
  Zhejiang University\\
  Hangzhou, China \\
  \texttt{hal\_42@zju.edu.cn} \\
  \And
  Xiaojin Gong\\
  Zhejiang University\\
  Hangzhou, China \\
  \texttt{gongxj@zju.edu.cn} \\
}

\begin{document}

\maketitle

\begin{abstract}
This work presents a tuning-free semantic segmentation framework based on classifying SAM masks by CLIP, which is universally applicable to various types of supervision.
Initially, we utilize CLIP's zero-shot classification ability to generate pseudo-labels or perform open-vocabulary segmentation.
However, the misalignment between mask and CLIP text embeddings leads to suboptimal results. 
To address this issue, we propose discrimination-bias aligned CLIP to closely align mask and text embedding, offering an overhead-free performance gain. 
We then construct a global-local consistent classifier to classify SAM masks, which reveals the intrinsic structure of high-quality embeddings produced by DBA-CLIP and demonstrates robustness against noisy pseudo-labels.
Extensive experiments validate the efficiency and effectiveness of our method, and we achieve state-of-the-art (SOTA) or competitive performance across various datasets and supervision types.
\end{abstract}
\section{Introduction}
Traditional fully supervised semantic segmentation (FSSS) methods typically fine-tune an ImageNet~\cite{ImageNet} pretrained classification model with pixel-level labels.
To reduce or even eliminate annotation effort, tasks like semi-supervised semantic segmentation (SSS), weakly supervised semantic segmentation (WSSS), and open-vocabulary semantic segmentation (OVSS) have been proposed. These tasks often rely on domain-specific techniques to compensate for sparsity of supervision.
With the rise of foundational models~\cite{Radford2021CLIP,Kirillov2023SAM,dinov2,GroundingDINO}, some methods~\cite{SEPL,FMA-WSSS,Sun2023SAMwsss,Grounded-SAM} now leverage their powerful generalization capabilities for semantic segmentation.
Very recently, Blume~\etal~\cite{regionbased} propose using DINO v2~\cite{dinov2} or MaskCLIP~\cite{MaskClip} to map masks generated by SAM~\cite{Kirillov2023SAM} into embeddings. A linear probe is then trained to classify these masks and produce semantic segmentation results under full supervision. 
This approach requires no fine-tuning of the foundation model or learning auxiliary modules like visual prompts. Thus, it is highly efficient and described as \textit{tuning-free}. 
Similarly, our framework employs a frozen CLIP to classify SAM masks and is tuning-free.
However, our approach distinguishes itself by being \textit{universally} applicable to various types of supervision, including FSSS, SSS, WSSS, and OVSS. 
This versatility is achieved by leveraging CLIP's zero-shot classification ability to generate pseudo-labels or perform open-vocabulary segmentation, coupled with constructing a noise-robust classifier. 
In contrast, Blume~\etal~\cite{regionbased} mainly relies on DINO v2 without exploring the potential of the vision-language model, and its simple linear probe struggles with challenging sparse supervision.

Our baseline is built on MaskCLIP~\cite{MaskClip} and SAM~\cite{Kirillov2023SAM}. Initially, we generate SAM masks following~\cite{FMA-WSSS}, then apply average pooling on MaskCLIP outputs to get corresponding mask embeddings. We use CLIP text embeddings to classify these masks, where OVSS results can be directly obtained. By supplementing sparse supervision on the training set with classifying results, we create pseudo labels for training a linear probe.
The main bottleneck of our baseline is CLIP's poor zero-shot classification performance, leading to noisy pseudo-labels and suboptimal OVSS results. This is because CLIP is not pretrained for fine-grained representation, resulting in misalignment between text and mask embeddings. 
To address this bottleneck, we consider 1) improving mask embedding alignment with text and 2) developing a classifier robust against noisy pseudo-labels when training data is available. 
We first propose discrimination-bias aligned CLIP (DBA-CLIP), which integrates SAM mask into CLIP computation to produce highly text-aligned mask embeddings.
We also find that DBA-CLIP improves the quality of mask embeddings, giving them good intrinsic structure and inter-class distinguishability. However, this advantage is overshadowed by noisy pseudo-labels during linear probe training.
Moreover, storing all mask embeddings for the entire dataset is feasible since masks are much fewer than pixels. 
Inspired by this, we develop a global-local consistent classifier (GLCC) based on label propagation~\cite{Label-Propagation}. GLCC predicts labels by ensuring global-local consistency and excels at handling training data with high-quality embeddings but noisy labels. In summary, the contributions of our work are as follows: 
\vspace{-0.25em}
\begin{itemize}[leftmargin=2em]
\vspace{-0.25em}
\item We extend the semantic segmentation framework based on SAM mask classification to various supervision types while maintaining its tuning-free advantage.
\vspace{-0.25em}
\item We introduce DBA-CLIP, which generates high-quality mask embeddings that align closely with text embeddings, thereby enhancing CLIP's zero-shot classification with zero overhead.
\vspace{-0.25em}
\item We propose GLCC, an efficient classifier designed to be robust against noisy labels by uncovering the intrinsic structure of training data. It demonstrates high performance across various supervision tasks, especially in sparse supervision scenarios.
\vspace{-0.25em}
\item Experiments across FSSS, SSS, WSSS, and OVSS show that despite being free from tuning and post-processing, as well as being universal-purpose, our approach achieves SOTA or competitive results compared to methods that require fine-tuning, post-processing, or are task-specific.
\end{itemize}

\section{Related Work}
\paragraph{Adapting CLIP for Semantic Segmentation}
CLIP~\cite{Radford2021CLIP} has been widely adopted for semantic segmentation due to its strong generalization ability.
Open-vocabulary semantic segmentation methods, like~\cite{MaskClip,SCLIP,GEM,ClipSurgery,CLIP-DIY},
obtain coarse pixel-level prediction from a frozen CLIP directly. 
A series of methods, including ours, can be summarized as generating class-agnostic masks before using CLIP to classify them.
However, obtaining text-aligned mask embeddings is challenging as CLIP is pretrained on the image level. Pooling on MaskCLIP~\cite{TAG,regionbased} or masking input images~\cite{han2023global,liang2023open} can help but not fully resolve this issue.
Fine-tuning CLIP~\cite{wu2023clipself,luo2023segclip} to improve its fine-grained representation is effective but costly.
Some methods train a mask proposal network~\cite{liu2023open,xu2023side,liang2023open,ding2022open,DeOP} or use SAM~\cite{Kirillov2023SAM} to generate masks~\cite{vs2024possam,jiao2023learning}, then fine-tune CLIP or train auxiliary modules to gather mask embeddings aligned to text.
In contrast, we slightly modify CLIP to make it mask-aware, achieving highly text-aligned mask embeddings without any overhead.

\paragraph{Label Propagation}
Label propagation~\cite{Label-Propagation} is initially proposed for transductive semi-supervised learning, which propagates label information from labeled to unlabeled data via nearest neighbor connections. 
Iscen~\etal~\cite{LP-DSSL} employs label propagation to create pseudo-labels and trains a CNN for semi-supervised image classification.
ReCLIP~\cite{ReCLIP} propagates labels on CLIP visual and text embeddings, generating pseudo-labels to adapt CLIP for classification. 
Label propagation can also be used to predict labels for new unseen data, known as inductive learning.
Douze~\etal~\cite{low-shot} classifies test images by performing online label propagation on a large image dataset, which is very costly. 
A contemporaneous work, ZLaP~\cite{ZLaP}, adopts an more efficient form of inductive label propagation similar to~\cite{LP-LinearN}, and uses it for zero-shot image classification. However, their form is basically a similarity-weighted kNN.
In this work, we derive a different form of inductive label propagation that is as efficient as ZLaP but yields better performance and apply it to semantic segmentation.

\section{Preliminary}
\label{sec:LP}
As a preliminary, we first introduce using transductive label propagation to smooth soft labels.
Given L2 normalized embeddings $X = [X_1, X_2, \cdots, X_N] \in \mathbb{R}^{N \times d}$ of $N$ samples and their corresponding soft label $P^c \in \mathbb{R}^{N \times K}$ of $K$ classes, smoothed soft label $P^*  \in \mathbb{R}^{N \times K}$ can be retrieved by label propagation~\cite{Label-Propagation}.
We start with constructing a zero-diagonal affinity matrix $S \in \mathbb{R}^{N \times N}$. $S$ is sparse, and $S_{ij}$ only equals to $X_i^T X_j$ when $X_j \in \mathrm{NN}_k(X_i)$, where $\mathrm{NN}_k(\cdot)$ finds $k$-nearest neighbors of $X_i$ in $X$. $S$ is then symmetrized and degree-normalized to $\mathcal{S} = D^{-\frac{1}{2}} (S + S^T) D^{-\frac{1}{2}}$, where $D \in \mathbb{R}^{N \times N} = \mathrm{diag}((S + S^T) \mathbf{1})$. Finally, the improved soft label $P^*$ can be obtained by solving the following linear systems with conjugate-gradient method:
\begin{equation}
\label{eq:LP}
    (\mathbf{I} - \alpha \mathcal{S}) P^* = P^c,
\end{equation}
where $\alpha$ is the hyper-parameter controlling propagation magnitude.

\section{The Proposed Method}
\begin{figure}
    \centering
    \includegraphics[width=1.0\linewidth]{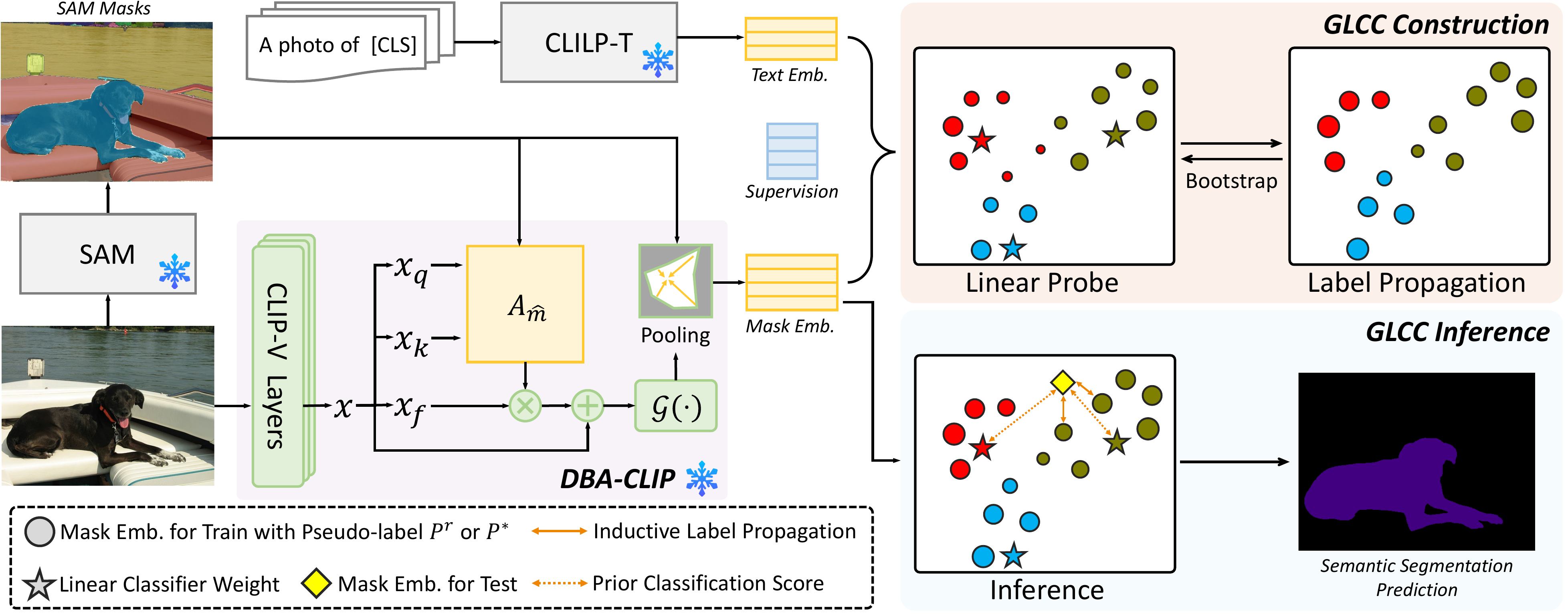}
    \caption{Method overview.
    The class and confidence of pseudo-labels are indicated by the color and size of the circles.
    DBA-CLIP is forwarded with awareness of SAM masks. Mask-pooling the DBA-CLIP output produces highly text-aligned mask embeddings, which, along with text embeddings and supervision, are used to construct GLCC. The construction involves alternating between training a linear probe and performing transductive label propagation. During inference, the GLCC classifies masks from test images to produce segmentation results.}
    \label{fig:overview}
\vspace{-1.5em}
\end{figure}
As shown in Figure~\ref{fig:overview}, our framework uses DBA-CLIP to extract embeddings for SAM masks before GLCC classifies them.
Specifically, for an input image $I$ and its corresponding SAM masks $\{m_u\}_{u=1}^{M}$, where $m_u \in \mathbb{B}^{H \times W}$ is a binary mask matching the image size, DBA-CLIP maps each $m_u$ into a mask embedding $e_u \in \mathbb{R}^d$. 
This $e_u$ is expected to match the text embedding from the CLIP text encoder, which encodes the class prompt of the image region covered by $m_u$.
Thus, the CLIP's zero-shot classifying results can be directly computed by $W e_u$, where $W \in \mathbb{R}^{K \times d}$ are normalized text embeddings for all classes.
These classifying results can be directly used for OVSS.
When supervision $Y \in \mathbb{R}^{N \times K}$ is available, all mask embeddings from training images are stacked into $X \in \mathbb{R}^{N \times d}$. 
We supplement the supervision $Y$ with zero-shot classification scores $X W^T$ to produce soft pseudo-labels $P^c$, upon which we construct our GLCC.
Methods for obtaining $Y$ and $P^c$ vary with supervision types, as detailed in Section~\ref{sec:GLCC}
During inference, mask embeddings for test images are classified by the constructed GLCC, producing semantic segmentation prediction.
Our method is tuning-free and avoids gradient backpropagation among backbone, eliminating time and space bottlenecks caused by fine-tuning. Additionally, both the construction and inference of GLCC are highly efficient. 

\subsection{Discrimination-Bias Aligned CLIP}
\label{sec:DBA-CLIP}

\begin{figure}
    \centering
    \includegraphics[width=1.0\linewidth]{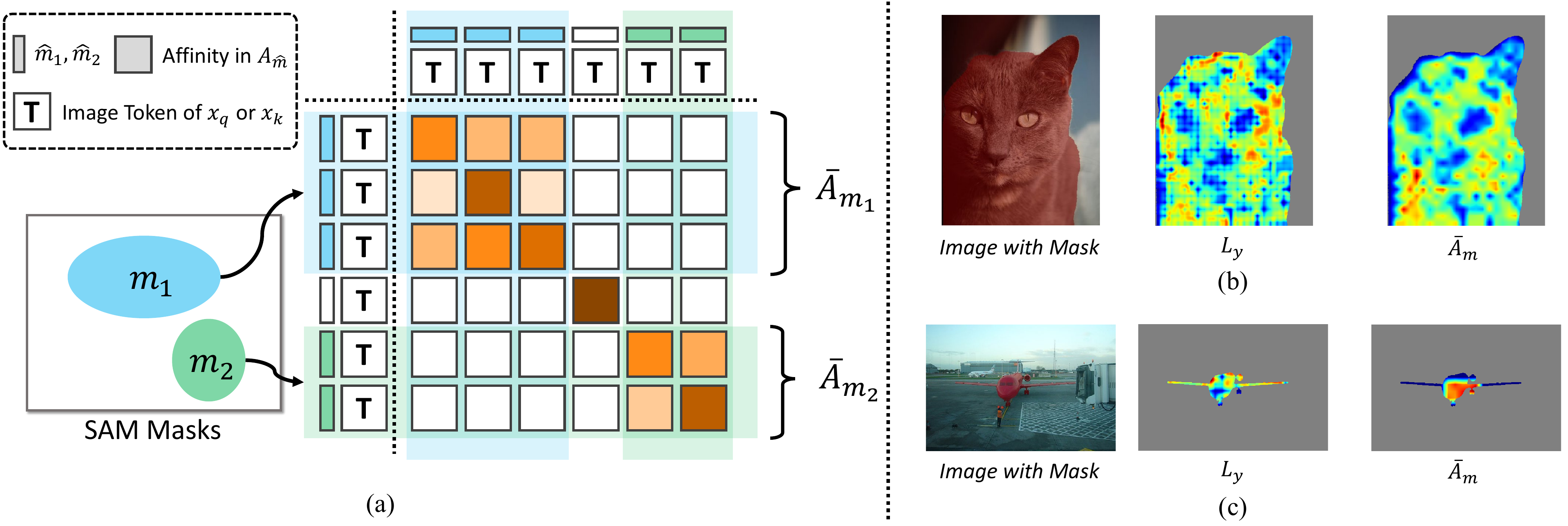}
    \caption{DBA-CLIP Illustration. (a) We generate attention bias based on masks to constrain affinity computation within the mask. (b) Affinities of the same mask are averaged to produce $\bar{A}_m$, which should align with the discrimination bias on the $L_y$. (c) However, the low-resolution $\bar{A}_m$ fails to match the contour of small and irregular objects, which leads us to adopt an approximate version that avoids directly using $\bar{A}_m$ to weighted-pool MaskCLIP output. See more examples in \ref{fig:More-DBA}.}
    \label{fig:DBA-CLIP}
\vspace{-1em}
\end{figure}

The naïve way to obtain mask embedding is to reshape and upsample the CLIP output to the original image size, then perform average pooling over the mask. Let us start by processing a single mask of class $y$, reshaped into a vector $m \in \mathbb{B}^{HW}$. The zero-shot classification score on class $y$, denoted as $l_y$, is computed by the cosine similarity between the mask embedding $e \in \mathbb{R}^{d}$ and the text embedding $W_y \in \mathbb{R}^{d}$. That is:
\begin{equation}
\label{eq:cls-score}
    l_y =  e^T W_y
\end{equation}
\begin{equation}
\label{eq:CLIP+SAMQ}
    e = \mathcal{N}( \frac{1}{|m|} m^T \mathcal{G}(x + A x_f)),
\end{equation}
\begin{equation}
\label{eq:CLIP+SAMQ.Aqkv}
    A = \mathrm{softmax}(\frac{x_q x_k^T}{d'}), x_q=\mathcal{W}_q(\mathrm{ln}(x)), x_k=\mathcal{W}_k(\mathrm{ln}(x)), x_f = \mathcal{W}_p(\mathcal{W}_v(\mathrm{ln}(x)),
\end{equation}
where $x \in \mathbb{R}^{l \times d'}$ are $l$ image tokens input to the last CLIP layer, and $\mathcal{G(\cdot)}$ denotes the final operations of CLIP, involving MLP, layer normalization, linear projection, upsampling and reshaping, resulting in output sized $HW \times d$.  $|m|$ is the mask area, $\mathcal{N}(\cdot)$ is L2 normalization on the last dimension, $\mathrm{ln}(\cdot)$ is layer normalization, and $\mathcal{W}(\cdot)$ denotes linear projections of the self-attention\footnote{We simplify the multi-head attention to single-head, and experiments show they are equally effective.}.
However, since only the class token is supervised during pretraining, the attention maps between image tokens in the last CLIP layer, that is $A \in \mathbb{R}^{l \times l}$ of Equation~\ref{eq:CLIP+SAMQ.Aqkv}, are meaningless and will disrupt the spatial correspondence. 
MaskCLIP~\cite{MaskClip} suggests removing this attention map and retaining only token-wise operations, leading to MaskCLIP output with correct localization:
\begin{equation}
\label{eq:MaskCLIP+SAMQ}
    e = \mathcal{N}( \frac{1}{|m|} m^T \mathcal{G}(x + x_f)),
\end{equation}
which serves for our baseline.
We illustrate that the baseline model is still suboptimal by examining the similarity $L_y$ between the text embedding and MaskCLIP output within the mask:
\begin{equation}
\label{eq:CAM}
    L_y = \mathcal{N}(\mathcal{G}(x + x_f)) W_y \odot m,
\end{equation}
where $\odot$ denotes Hadamard product, and $L_y$ is reshaped to $H \times W$ for visualization in Figure~\ref{fig:DBA-CLIP}.
It can be found that $L_y$ is only strongly activated in parts of the object, diminishing the mask classification score $l_y$. This phenomenon, termed \textbf{discrimination-bias} (DB), originates from pretraining CLIP at the image level, leading it to classify through the most discriminative parts.
To compensate the drop in classification scores, one strategy is to re-weight MaskCLIP output according to DB, \textit{i.e.}\ increasing the contribution of discriminative parts to the mask embedding. The challenge of this strategy lies in determining the distribution of DB without knowing class $y$. Fortunately, we find that the average affinity among image tokens within the mask, denoted as $\bar{A}_{m} \in \mathbb{R}^{HW}$, matches DB closely and can serve as a substitute, as Figure~\ref{fig:DBA-CLIP} (b) illustrates. Thus, we leverage the average affinity $\bar{A}_{m}$ to align MaskCLIP output weights to the DB by:
\begin{equation}
\label{eq:DBA-CLIP-bad+SAMQ}
    e = \mathcal{N}(\bar{A}_{m}^T \mathcal{G}(x + x_f)),
\end{equation}
\begin{equation}
\label{eq:DBA-CLIP-bad+SAMQ.A_bar}
\bar{A}_{m} = \mathcal{U}(\frac{1}{|\hat{m}|} \hat{m}^T A_{\hat{m}}) \odot m,
\end{equation}
\begin{equation}
\label{eq:DBA-CLIP+SAMQ.A_m}
A_{\hat{m}} = \frac{1}{2} (\mathrm{softmax}(\frac{x_q x_q^T}{d'} - c (1 - B)) + \mathrm{softmax}(\frac{x_k x_k^T}{d'} - c (1 - B)))), \text{where} \ B = \hat{m} \hat{m}^T \vee \mathbf{I},
\end{equation}
where $c$ is a large constant, $\hat{m} \in \mathbb{B}^{l}$ is the downsampled mask, and $\mathcal{U}(\cdot)$ upsamples and reshapes the input to $HW$. The affinities among image tokens are measured by the intra-similarity of $x_q$ and $x_k$, as SCLIP~\cite{SCLIP} suggests.
Eventually, we adopt an approximate version of Equation~\ref{eq:DBA-CLIP-bad+SAMQ} (see proof in Appendix~\ref{sec:proof-DAB-CLIP}) as the final implementation of DBA-CLIP:
\begin{equation}
\label{eq:DBA-CLIP+SAMQ}
    e = \mathcal{N}( \frac{1}{|m|} m^T \mathcal{G}(x + A_{\hat{m}} x_f)).
\end{equation}
We adopt this approximate version for two main reasons: 1) It can be very easily implemented by adding an attention bias. 2) As shown in Figure~\ref{fig:DBA-CLIP} (c), $\bar{A}_{m}$ is upsampled from a lower size, inevitably missing detailed object contours compared to the mask and consequently overlooking prominent or elongated parts of objects. Equation~\ref{eq:DBA-CLIP+SAMQ} does not explicitly compute $\bar{A}_{m}$ nor uses it to re-weight MaskCLIP output, thus avoiding this issue and enhancing performance.

DBA-CLIP may bear some similarity to key smoothing~\cite{MaskClip} or SCLIP~\cite{SCLIP}, but they are fundamentally different.
These methods aim to locate entire objects, including non-discriminative regions, which is not very helpful for us since SAM masks already identify the complete object. Our challenge shift to correctly classify these masks.
Thus, we compute in a mask-aware manner, obtaining more text-aligned mask embeddings than methods like SCLIP by aligning DB.
Finally, the MPS of DeOP~\cite{DeOP} tries to constrain the attention between $x_q$ and $x_k$, that is $A$ of Equation~\ref{eq:CLIP+SAMQ.Aqkv}, within masks for a CLIP with learnable visual prompts. However, as noted, this attention is meaningless and not worth further exploration. Also, that is why DeOP ultimately turns back to MaskCLIP~\cite{MaskClip}, \textit{i.e.}, Equation~\ref{eq:MaskCLIP+SAMQ}.

\subsection{Global-Local Consistent Classifier}
\label{sec:GLCC}
On the training set, given high-quality mask embeddings $X$ produced by DBA-CLIP and soft pseudo-labels $P^c \in \mathbb{R}^{N \times K}$, which are noisy under weak or semi-supervision, there are two approaches to build the classifier efficiently: linear probe and inductive label propagation. We let them mutually bootstrap to obtain GLCC. 

\paragraph{Obtaining Soft Pseudo-label} 
We adopt the soft label for it is confidence-aware. For full supervision, the supervision $Y$ is the proportion of pixels for each class within masks. We always let pseudo-label $P^c = Y$. 
When semi-supervised, the values of $Y$ for labeled samples remain consistent with full supervision. We set $P^c$ to $Y$ on its labeled part and fill the remaining unlabeled part with CLIP's zero-shot classification scores $\mathrm{softmax}(X W^T)$.
The weakly supervised $Y$ is the multi-class label of the image containing the mask. We set $P^c = \mathrm{softmax}(X W^T \odot Y - c(1 - Y))$, which filters the zero-shot classification scores with image-level labels.
GLCC is not applicable to the OVSS.

\paragraph{Linear Probe}
We trained a linear classifier $U \in \mathbb{R}^{K \times d}$ on $X$ supervised by $P^c$, employing cross-entropy loss weighted by the mask areas.
$U e^v$ computes the classifying result for the mask embedding $e^v \in \mathbb{R}^d$ from the test set. 
Besides, supplementing the supervision $Y$ with classification scores $\mathrm{softmax}(X U^T)$ of linear probe produces another set of pseudo-label $P^r$.

\paragraph{Inductive Label Propagation} 
Label propagation~\cite{Label-Propagation} exploits the intrinsic structure of the embeddings and smooths soft labels by enforcing \textit{global-local consistency}, aiming for consistent label assignment among embeddings neighbored (local consistency) or within the same structure (global consistency).
Inspired by this, we can develop a better classifier optimized for learning from high-quality yet noisily labeled embeddings.
The original label propagation is transductive and only smooths pseudo-labels within the training set. By allowing the test sample $e^v$ and its prior classification score $p \in \mathbb{R}^{K}$, which can be set to all zeros if unknown, to participate in the propagation as well, the resulting $p^*$ can serve as the classifying result for $e^v$. That is:
\begin{equation}
\label{eq:GLCC-bad}
(\mathbf{I} - \alpha \bar{\mathcal{S}}) 
\begin{bmatrix}
\bar{P}^*\\
p^*
\end{bmatrix}
= 
\begin{bmatrix}
P^c\\
p
\end{bmatrix},
\end{equation}
\begin{equation}
\label{eq:GLCC-bad.S_bar}
\bar{\mathcal{S}} = \bar{D}^{-\frac{1}{2}} (\bar{S} + \bar{S}^T) \bar{D}^{-\frac{1}{2}}, \ \text{where} \ \bar{D} = \mathrm{diag}((\bar{S} + \bar{S}^T) \mathbf{1}),
\end{equation}
\begin{equation}
\label{eq:GLCC-bad.S}
\bar{S}_{i j} = 
\begin{cases}
\bar{X}_i^T \bar{X}_j, \ \text{if} \ i \neq j \wedge \bar{X}_j \in \mathrm{NN}_k(\bar{X}_i) \\
0, \text { otherwise },
\end{cases}, \ \text{where} \
\bar{X} =
\begin{bmatrix}
X\\
e^v
\end{bmatrix}.
\end{equation}
However, solving Equation~\ref{eq:GLCC-bad} for each $p^*$ is unacceptably expensive. Thus, we adopt an approximate solution (see proof in Appendix~\ref{sec:proof-GLCC}) with $O[d \sqrt{N}]$ complexity~\cite{douze2024faiss}:
\begin{equation}
\label{eq:GLCC}
    p^* = p + \alpha \bar{a}^T P^*
\end{equation}
\begin{equation}
\label{eq:GLCC.a}
\bar{a}_i = 
\frac{a_i}{\sqrt{\frac{1}{2} D_{i i} \sum_{i=1}^{N} a_i}}, \ \text{where} \
a_i = 
\begin{cases}
X_i^T e^v, \ \text{if} \ X_i \in \mathrm{NN}_{k}(e^v) \\
0, \text { otherwise },
\end{cases}.
\end{equation}
Here, we first perform transductive label propagation on the training set with $X$ and $P^c$ following Section~\ref{sec:LP}, saving the degree matrix $D$ and the smoothed pseudo-label $P^*$ for later use.
During inference, for each $e^v$, we select $k$-nearest neighbors from $X$ and aggregate their corresponding smoothed labels in $P^*$ with weights $\bar{a} \in \mathbb{R}^{N}$. The aggregated result is modulated by $\alpha$ and used to revise the prior classification score $p$. 
Previous works~\cite{LP-LinearN,ZLaP} also explore the inductive usage of label propagation, but they ultimately lead to a similarity-weighted kNN. Our method differs in that: 1) Incorporates prior classification score $p$. 2) Applies weights $\bar{a}$ with the degree-normalization term $D_{ii}$ to reduce the impact of neighbors in dense sub-clusters and prevent them from dominating the classifying result.

\paragraph{Mutually Bootstrap} 
Our GLCC first learns a linear probe $U$ on $P^c$ and obtains improved $P^r$ when $P^c$ is noisy. 
We then run transductive label propagation on $P^r$ instead of $P^c$ to construct inductive label propagation. During inference, classifying results of the linear probe are used as prior classification score in Equation~\ref{eq:GLCC}, \textit{i.e.}, $p = \mathrm{softmax}(U e^v)$. 
Finally, supplementing the supervision $Y$ with transductive label propagation results $P^*$ produces a new $P^c$, upon which another round of bootstrap can be performed.
Note that for FSSS, we reset $P^*$ to $Y$ and only perform one round of bootstrap.

\section{Experiments}
\subsection{Datasets and Evaluation Metrics}
\label{sec:datasets}
We evaluate our method on five widely used datasets. 
\textbf{PASCAL VOC 2012}~\cite{everingham2010pascal} has 20 object classes and one background class, containing 1,464, 1,449 and 1,456 images for training, validation and testing. Following~\cite{SBD}, an augmented set with 10,582 images is used for training. We experiment with all types of supervision on PASCAL VOC 2012. For OVSS, we evaluate it under two settings: with and without background, denoted as VOC21 and VOC20.
\textbf{COCO-Obj}~\cite{COCO17} includes 118,287 training and 5,000 validation images, with 80 object classes and one background class. We experiment with COCO-Obj for FSSS, SSS and OVSS. For SSS, we use the same random partition as UniMatch~\cite{UniMatch}. 
\textbf{MS COCO 2014}~\cite{Lin2014COCO} is the subset of COCO-Obj with a different split, containing 82,081 images for training and 40,137 images for validation. We experiment with it for WSSS. 
\textbf{COCO-Stuff}~\cite{COCO17} shares the same data and split with COCO-Obj but is annotated with 171 semantic classes without background class.
\textbf{Cityscapes}~\cite{cordts2016cityscapes} consists of 19 semantic classes and has no background class, including 500 validating images with fine annotations. The COCO-Stuff and Cityscapes are only evaluated for OVSS.
In addition to the widely used semantic segmentation metric mIoU, we also employ the F$_1$ score to evaluate the classification performance on SAM masks.

\subsection{Implementation Details}
\label{sec:impl}
We generate minimal-overlapping SAM masks following FMA-WSSS~\cite{FMA-WSSS}. 
The ViT-B/16~\cite{Dosovitskiy2021} image encoder and the transformer text encoder pretrained by CLIP~\cite{Radford2021CLIP} are adopted.
When inferring DBA-CLIP, the entire image is first input to obtain global output. Next, we resize the input image with a short side of 448 and perform slide inference with a 224×224 window and 112 stride. We align the scale and position of both global and sliding window output, calculate their average, and pool on this average output to produce the final mask embedding. We denote this strategy as global-aware sliding window (GSW).
The hyper-parameters for linear probe training are set following MAE~\cite{MAE}. Unlike MAE, since masks are much fewer than pixels, we pre-store mask embeddings of training set instead of inferring them in real-time, thereby significantly accelerating the linear probe training.
We set $k = 50$ and $\alpha = 0.9$ for inductive label propagation.
Our method is implemented with PyTorch~\cite{paszke2019pytorch} and faiss~\cite{douze2024faiss, gpu-faiss}. All experiments are conducted on a single RTX 3090 GPU with 24GB memory. 

\begin{table}[htpb]
\centering
\caption{
Fully supervised semantic segmentation results on PASCAL VOC 2012 and COCO-Obj.
The \ddag~indicates backbone pretrained on ImageNet-21k~\cite{ImageNet-21k}. "\textit{T.F.}" indicates whether it is tuning-free.
}
\label{tab:FSSS-SOTA}
\scalebox{1.0}{
\begin{tabular}{llcccc}
\hline
\multirow{2}{*}{Methods} & \multirow{2}{*}{Backbone} & \multirow{2}{*}{\textit{T.F.}} & \multicolumn{2}{c}{VOC} & COCO-Obj \\
\cmidrule(lr){4-5} \cmidrule(lr){6-6} 
& & & \textit{val} & \textit{test} & \textit{val} \\
\hline
DeepLab V3+{$_{\text{\color{gray}{CVPR18}}}$}~\cite{Chen2018deeplab} & R101 & \XSolidBrush & 79.5 & 79.4 & 55.4\\

Mask2Former{$_{\text{\color{gray}{CVPR22}}}$}~\cite{mask2former} & Swin-L\ddag & \XSolidBrush & 86.0 & 86.1 & 64.0\\

Blume~\etal{$_{\text{\color{gray}{CVPR24}}}$}~\cite{regionbased} & SAM + CLIP & \Checkmark & 76.7 & - & - \\

Blume~\etal{$_{\text{\color{gray}{CVPR24}}}$}~\cite{regionbased} & SAM + DINO v2 & \Checkmark & 86.9 & - & - \\

\rowcolor{gray!40} \textbf{Ours} & SAM + CLIP & \Checkmark & 84.2 & 83.9 & 59.5\\
\hline
\end{tabular}
}
\vspace{-1em}
\end{table}
\begin{table}[htpb]
\caption{Semi-supervised semantic segmentation results on PASCAL VOC 2012 and COCO-Obj. Best results are marked in bold.}
\centering
\label{tab:SSS-SOTA}
\resizebox{1.0\textwidth}{!}{
\begin{tabular}{lcccccccccc}
\hline
\multirow{3}{*}{Methods} & \multirow{3}{*}{\textit{T.F.}} & \multicolumn{4}{c}{VOC} & \multicolumn{5}{c}{COCO-Obj} \\
\cmidrule(lr){3-6} \cmidrule(lr){7-11}
 & & 1/16 & 1/8 & 1/4 & 1/2 & 1/512 & 1/256 & 1/128 & 1/64 & 1/32 \\
 & & (662) &  (1323) &  (2646) &  (5291) &  (232) & (463) &  (925)  &  (1849) & (3697) \\
\hline
PseudoSeg{$_{\text{\color{gray}{ICLR21}}}$}~\cite{PsudoSeg} & \XSolidBrush & - & - & - & - & 29.8 & 37.1 & 39.1 & 41.8 & 43.5 \\ 

PC$^2$Seg{$_{\text{\color{gray}{ICCV21}}}$}~\cite{PC2Seg} & \XSolidBrush & - & - & - & - & 29.9 & 37.5 & 40.1 & 43.7 & 46.1 \\

FPL{$_{\text{\color{gray}{CRPR23}}}$}~\cite{FPL} & \XSolidBrush & 75.0 & 77.8 & 78.3 & - & - & - & - & - & - \\

AugSeg{$_{\text{\color{gray}{CRPR23}}}$}~\cite{AugSeg}  & \XSolidBrush & 77.0 & 77.3 & 78.8 & - & - & - & - & - & - \\

DGCL{$_{\text{\color{gray}{CRPR23}}}$}~\cite{DGCL} & \XSolidBrush & 76.6 & 78.4 & 79.3 & 81.0 & - & - & - & - & - \\

UniMatch{$_{\text{\color{gray}{CRPR23}}}$}~\cite{UniMatch} & \XSolidBrush & 78.1 & 78.4 & 79.2 & - & 31.9 & 38.9 & 44.3 & 48.2 & 49.8 \\

ESL{$_{\text{\color{gray}{ICCV23}}}$}~\cite{UniMatch} & \XSolidBrush & 76.4 & 78.6 & 79.0 & 80.0 & - & - & - & - & - \\

CFCG{$_{\text{\color{gray}{ICCV23}}}$}~\cite{CFCG} & \XSolidBrush & 76.8 & 79.1 & 80.0 & 80.2 & - & - & - & - & - \\

DLG{$_{\text{\color{gray}{ICCV23}}}$}~\cite{DLG} & \XSolidBrush & 77.8 & 79.3 & 79.1 & 79.5 & - & - & - & - & - \\

LogicDiag{$_{\text{\color{gray}{ICCV23}}}$}~\cite{LogicDiag}  & \XSolidBrush & - & - & - & - & 33.1 & 40.3 & 45.4 & 48.8 & 50.5 \\

MKD{$_{\text{\color{gray}{MM23}}}$}~\cite{MKD}  & \XSolidBrush & - & - & - & - & 36.7 & 43.7 & 48.9 & \textbf{51.0} & \textbf{54.1} \\

AllSpark{$_{\text{\color{gray}{CVPR24}}}$}~\cite{AllSpark}  & \XSolidBrush & 78.3 & 80.0 & 80.4 & 81.1 & 34.1 & 41.6 & 45.5 & 49.6 & - \\

\rowcolor{gray!40} \textbf{Ours} & \Checkmark & \textbf{81.5} & \textbf{82.5} & \textbf{83.1} & \textbf{83.5} & \textbf{48.8} & \textbf{49.6} & \textbf{50.1} & 50.6 & 51.4 \\
\hline
\end{tabular}
}
\vspace{-1em}
\end{table}

\begin{table}[t]
\centering
\caption{
Single-stage weakly supervised semantic segmentation results on PASCAL VOC 2012 and MS COCO 2014.
The type of supervision is denoted in the "Sup." column, including image-level labels ($\mathcal{I}$), using CLIP ($\mathcal{C}$) and SAM ($\mathcal{S}$). The \ddag~means ImageNet-21k~\cite{ImageNet-21k} pretrained backbone. Refer to Table~\ref{tab:WSSS-SOTA-MS} for multi-stage results.
}
\label{tab:WSSS-SOTA-SS}
\begin{tabular}{lcclccc}
\toprule
\multirow{2}{*}{Methods} & \multirow{2}{*}{\textit{T.F.}} & \multirow{2}{*}{\textit{Sup.}} & \multirow{2}{*}{Backbone} & \multicolumn{2}{c}{VOC} & COCO \\
\cmidrule(lr){5-6} \cmidrule(lr){7-7}
& & & & \textit{val} & \textit{test} & \textit{val} \\
\midrule

\multicolumn{6}{l}{\textit{\textbf{Single-stage (End-to-end) Methods}}} \\

1Stage{$_{\text{\color{gray}{CVPR20}}}$}~\cite{1Stage} & \XSolidBrush  & $\mathcal{I}$ & WR38 & 62.7 & 64.3 & - \\

RRM{$_{\text{\color{gray}{AAAI20}}}$}~\cite{RRM} & \XSolidBrush  & $\mathcal{I}$ & ViT-B & 63.1 & 62.4 & - \\

AFA{$_{\text{\color{gray}{CVPR22}}}$}~\cite{AFA} & \XSolidBrush  & $\mathcal{I}$ & MiT-B1 & 66.0 & 66.3 & 38.9 \\

SLRNet{$_{\text{\color{gray}{IJCV22}}}$}~\cite{SLRNet} & \XSolidBrush  & $\mathcal{I}$ & WR38 & 67.2 & 67.6 & 35.0 \\

ViT-PCM{$_{\text{\color{gray}{ECCV22}}}$}~\cite{ViT-PCM} & \XSolidBrush & $\mathcal{I}$ & ViT-B\ddag & 70.3 & 70.9 & 45.0 \\

ToCo{$_{\text{\color{gray}{CVPR23}}}$}~\cite{ToCo} & \XSolidBrush & $\mathcal{I}$ & ViT-B\ddag & 71.1 & 72.2 & 42.3 \\

DuPL{$_{\text{\color{gray}{CVPR24}}}$}~\cite{DUPL} & \XSolidBrush & $\mathcal{I}$ & ViT-B\ddag & 73.3 & 72.8 & 44.6 \\

SeCo{$_{\text{\color{gray}{CVPR24}}}$}~\cite{SeCo} & \XSolidBrush & $\mathcal{I}$ & ViT-B\ddag & 74.0 & 73.8 & 46.7 \\

CoSA{$_{\text{\color{gray}{arXiv24}}}$}~\cite{CoSA} & \XSolidBrush  & $\mathcal{I}$ & ViT-B\ddag & 76.4 & 75.2 & 51.1 \\

\rowcolor{gray!40} \textbf{Ours} & \Checkmark  &  $\mathcal{I} + \mathcal{C} + \mathcal{S}$  & SAM + CLIP & \textbf{81.8} & \textbf{81.1} & \textbf{53.5}\\ 

\bottomrule
\end{tabular}
\vspace{-1em}
\end{table}
\begin{table}[t]
\centering
\caption{Training-free open-vocabulary semantic results on five datasets. \textit{Post.} denote applying a post-processing like CRF~\cite{Krahenbuhl2011CRF} or PAMR~\cite{PAMR} which are computationally expensive.}
\label{tab:OV-SOTA}
\resizebox{1.0\textwidth}{!}{%
\begin{tabular}{lcccccc}
\toprule
\multirow{2}{*}{Methods} & \multirow{2}{*}{\textit{Post.}} & \multicolumn{2}{c}{\textit{With a background category}} & \multicolumn{3}{c}{\textit{Without background category}} \\
\cmidrule(r){3-4} \cmidrule(r){5-7}
 & & VOC21 & COCO-Obj & VOC20 & CityScapes & COCO-Stuff \\
\midrule
CLIP{$_{\text{\color{gray}{ICML21}}}$}~\cite{Radford2021CLIP} & \Checkmark & 19.8 & 10.4 & 54.2 & 7.0 & 5.9 \\

MaskCLIP{$_{\text{\color{gray}{ECCV22}}}$}~\cite{MaskClip} & \Checkmark & 52.0 & 22.6 & 72.1 & 30.1 & 20.0\\

GroupViT{$_{\text{\color{gray}{CVPR22}}}$}~\cite{GroupVit} & \Checkmark & 52.7 & 27.9 & 81.5 & 21.7 & 16.9\\

ReCo{$_{\text{\color{gray}{NIPS22}}}$}~\cite{ReCo} & \Checkmark & 27.2 & 17.3 & 62.4 & 23.2 & 16.3\\

TCL{$_{\text{\color{gray}{CVPR23}}}$}~\cite{TCL} & \Checkmark & 55.0 & 31.6 & 83.2 & 24.3 & 22.4\\

CLIP-DIY{$_{\text{\color{gray}{WACV24}}}$}~\cite{CLIP-DIY} & \XSolidBrush & 59.9 & 31.0 & - & - & - \\

FOSSIL{$_{\text{\color{gray}{WACV24}}}$}~\cite{FOSSIL} & \XSolidBrush & - & - & - & 23.2 & 24.8 \\

SCLIP{$_{\text{\color{gray}{arXiv23}}}$}~\cite{SCLIP} & \Checkmark & 61.7 & 32.1 & 83.5 & 34.1 & 23.9 \\

\rowcolor{gray!40} \textbf{Ours} & \XSolidBrush & \textbf{74.3} & \textbf{43.8} & \textbf{86.7} & \textbf{44.1} & \textbf{29.3} \\
\bottomrule
\end{tabular}%
}
\vspace{-1em}
\end{table}
\begin{table}[htpb]
\centering
\caption{
Performance comparison of the proposed method and its variants on the PASCAL VOC 2012 \textit{val} set under different types of supervision. For semi-supervision, we experiment on the 1/16 (662) partition.
}
\label{tab:Main-Abl}
\resizebox{1.0\textwidth}{!}{
\begin{tabular}{lcccccccc}
\hline
\multirow{2}{*}{Methods} & \multicolumn{2}{c}{\textit{Fully Supervised}} & \multicolumn{2}{c}{\textit{Semi-Supervised}} & \multicolumn{2}{c}{\textit{Weakly Supervised}} & \multicolumn{2}{c}{\textit{Open-Vocabulary}} \\
\cmidrule(r){2-3} \cmidrule(r){4-5} \cmidrule(r){6-7} \cmidrule(r){8-9}
 & F$_1$(\%) & mIoU(\%) & F$_1$(\%) & mIoU(\%) & F$_1$(\%) & mIoU(\%) & F$_1$(\%) & mIoU(\%) \\
\hline
Baseline & 72.2 & 81.9 & 66.8 & 75.4 & 67.0 & 75.5 & 58.3 & 68.5 \\

+ DBA-CLIP & 74.4 & 83.2 & 70.5 & 78.3 & 70.6 & 78.1 & 65.8 & 74.3 \\

+ DBA-CLIP + GLCC & 77.9 & 84.2 & 75.5 & 81.5 & 75.3 & 81.8 & - & - \\
\hline
\end{tabular}
}
\vspace{-1em}
\end{table}

\subsection{Comparison to State-of-the-Art}
\paragraph{Fully Supervised Semantic Segmentation} 

In Table~\ref{tab:FSSS-SOTA}, we compare the FSSS results with other methods. 
DeepLab V3+~\cite{Chen2018deeplab} with ResNet101~\cite{ResNet} backbone and Mask2Former~\cite{mask2former} with Swin-L~\cite{swin} backbone are widely used semantic segmentation methods based on CNN and transformer. Both methods require fine-tuning. Our tuning-free method consistently surpasses the CNN counterpart and performs comparably to the transformer-based method on VOC.
On the same SAM + CLIP backbone, we significantly outperform Blume~\etal~\cite{regionbased}, which is also tuning-free. However, it outperforms us with a SAM + DINO v2 backbone. We insist on CLIP backbone since CLIP's zero-shot classification ability is crucial in non-fully supervised scenarios.

\paragraph{Semi-Supervised Semantic Segmentation} 

Table~\ref{tab:SSS-SOTA} compares our semi-supervised semantic segmentation results with other leading methods, among which we are the only tuning-free method. We achieve state-of-the-art performance on all partitions of VOC and the three most challenging partitions of COCO-Obj. Due to increasing labeled samples being more beneficial for methods that allow fine-tuning, we ranked second on the last two partitions of COCO-Obj.

\paragraph{Weakly Supervised Semantic Segmentation}
The most effective and widely used weakly supervised semantic segmentation pipeline is multi-stage. It first generates pseudo-masks on the training set, then trains a standard semantic segmentation network for testing. In contrast, single-stage methods use the same framework throughout. Our method can be directly applied to the test set for single-stage results or generate pseudo-masks on the training set before training a segmentation network following~\cite{FMA-WSSS} for multi-stage results.
As shown in Table~\ref{tab:WSSS-SOTA-SS}, as the only tuning-free method, we significantly outperform all other single-stage counterparts. In Table~\ref{tab:WSSS-SOTA-MS}, we also establish a new state-of-the-art performance for multi-stage methods.
Besides, previous methods always heavily rely on offline refinement modules like PAMR~\cite{PAMR}, CRF~\cite{Krahenbuhl2011CRF} or AffinityNet~\cite{Ahn2018AffinityNet,Lin2023CLIP-ES}. Instead, we implement our methods without any offline refinement.

\paragraph{Open-Vocabulary Semantic Segmentation}
In Table~\ref{tab:OV-SOTA}, we evaluate our proposed method against other training-free open-vocabulary semantic segmentation methods. With no training data, we use CLIP text embeddings directly as the classifier instead of constructing GLCC. We achieve remarkable performance and significantly surpass previous SOTA methods without any post-processing.
\subsection{Ablation Studies}
\label{sec:ABL}
\paragraph{Effectiveness of Main Components}
Table~\ref{tab:Main-Abl} investigates the effective of our DBA-CLIP and GLCC on PASCAL VOC 2012 under all types of supervision. The baseline model average-pools on MaskCLIP~\cite{MaskClip} output (Equation~\ref{eq:MaskCLIP+SAMQ}) for mask embeddings and train a linear probe to classify them. 
It can be observed that sparser supervision leads to greater improvement of DBA-CLIP and GLCC. 
This is because DBA-CLIP closely aligns mask and text embeddings, fully leveraging CLIP’s zero-shot classification ability to address supervision sparsity. DBA-CLIP also reduces the impact of semantically poor regions and produces more distinctive embeddings, bringing improvements even under full supervision.
Moreover, the plain linear probe cannot distinguish noise in pseudo-labels. In contrast, our GLCC excels at revealing the intrinsic structure of high-quality embeddings, eliminating the noise, and achieving accurate classifying results.
Finnaly, the inductive label propagation of GLCC can better classify embeddings near decision boundaries, typically from small masks. 
This explains why GLCC improves F$_1$ scores more than mIoU.
We make a qualitative comparison under weak supervision in Figure~\ref{fig:Seg}.

\paragraph{Effectiveness of DBA-CLIP Designs}
Table~\ref{tab:DBA-Abl} evaluates the pseudo-labels obtained by DBA-CLIP and its counterparts under weak supervision, where the supervision exists but is very sparse, not overshadowing the advantages of DB alignment. This setup can typically demonstrate the effectiveness of DBA-CLIP.
As explained in Section~\ref{sec:DBA-CLIP}, DeOP with MPS is ineffective, and SCLIP is no more helpful than MaskCLIP. The slight improvement with SCLIP can be attributed to images where a single object occupies most of the area, making the effect of SCLIP similar to DBA-CLIP.
\begin{wraptable}{r}{0.5\textwidth}
\centering
\caption{
Performance comparison of the DBA-CLIP and its variants on the PASCAL VOC 2012 \textit{trainaug} set. "\textit{Down.}" indicates downsampling mask and pooling at a lower size. "GSW" denotes adopting global-aware sliding window startegy. "MaskCLIP" refers to pooling on its output to obtain SAM mask embeddings, the same for "SCLIP" and "DeOP".
}
\label{tab:DBA-Abl}
\resizebox{1.0\linewidth}{!}{
\begin{tabular}{lcc}
\hline
Methods & F$_1$(\%) & mIoU(\%) \\
\hline
MaskCLIP (Eq.~\ref{eq:MaskCLIP+SAMQ}) & 62.8 & 71.2 \\
MaskCLIP + GSW & 65.3 & 72.8 \\
SCLIP~\cite{SCLIP} & 63.5 & 71.5 \\
DeOP with MPS~\cite{DeOP} & 37.8 & 52.5 \\
Origin DBA-CLIP (Eq.~\ref{eq:DBA-CLIP-bad+SAMQ}) & 65.5 & 73.7 \\
Origin DBA-CLIP + \textit{Down.} & 63.6 & 73.4 \\
DBA-CLIP (Eq.~\ref{eq:DBA-CLIP+SAMQ}) & 66.7 & 74.3 \\
DBA-CLIP + GSW & 69.2 & 75.6 \\
\hline
\end{tabular}
}
\vspace{-1em}
\end{wraptable}
The origin DBA-CLIP shows great performance gain over MaskCLIP but struggles with mismatches between low-resolution average affinity and mask contours. In the sixth row, we attempt to downsample the mask and pool at a lower size, but leading to poorer performance due to the loss of contour details during downsampling.
The final DBA-CLIP avoids such issues and achieves slightly better mIoU with a simpler implementation. Additionally, since small masks are more sensitive to the mismatch issue, the improvement in F$_1$ is greater.
Finally, we find that our GSW strategy consistently boosts performance.
\paragraph{Effectiveness of GLCC Designs}
In Table~\ref{tab:GLCC-Abl}, we validate the effectiveness of each component of GLCC under weak supervision. 
It can be observed that when used alone, inductive label propagation outperforms the linear probe, especially in terms of F$_1$.
Using both together and bootstrapping each other brings greater improvement, whereas a single classifier bootstrapping itself offers minimal benefits. The order of bootstrapping does not matter.
Table~\ref{tab:IL-Abl} further compares different forms of inductive label propagation, where first row is the forms used in~\cite{LP-LinearN,ZLaP}. Incorporating prior classification scores and using degree-normalized weights derived from Sec~\ref{sec:proof-GLCC} instead of simple L1 normalization both lead to better performance. The improvement in F$_1$ is mainly due to better normalization weights. The fourth row is our final form, achieving the best performance.
Finally, Table~\ref{tab:ak-Abl} shows that GLCC is robust to the hyperparameter $k$, but $\alpha$ needs careful selection. The appendix~\ref{sec:GLCC-eff} also discusses the efficiency of GLCC.

\FloatBarrier
\begin{figure}[htpb]
\centering

\vspace{-0.5em}
\centering
\begin{minipage}{0.48\textwidth}
\captionof{table}{
Performance comparison of the GLCC and its variants on the PASCAL VOC 2012 \textit{val} set. "\textit{LP}" indicates linear probe and "\textit{IL}" indicates inductive label propagation. "$\to$" indicates using one classifier to bootstrap another one.
}
\label{tab:GLCC-Abl}
\begin{tabular}{lcc}
\hline
Methods & F$_1$(\%) & mIoU(\%) \\
\hline
\textit{LP} & 70.6 & 78.1 \\
\textit{IL} & 74.5 & 79.7 \\
\textit{LP} $\to$ \textit{IL} & 74.8 & 81.1 \\
\textit{LP} $\to$ \textit{LP} & 71.1 & 78.3\\
\textit{IL} ~$\to$ \textit{IL} & 74.7 & 79.8\\
\textit{LP} $\to$ \textit{IL} ~$\to$ \textit{LP} $\to$ \textit{IL} & 75.3 & 81.8 \\
\textit{IL} ~$\to$ \textit{LP} $\to$ \textit{IL} ~$\to$ \textit{LP} & 75.5  & 81.8 \\
\hline
\end{tabular}

\end{minipage}
\hfill
\centering
\begin{minipage}{0.48\textwidth}
\captionof{table}{
Performance comparison of the inductive label propagation and its variants on the PASCAL VOC 2012 \textit{val} set. "$\frac{a^T}{||a||}$" indicates L1 normalized similarities.
}
\label{tab:IL-Abl}
{\renewcommand{\arraystretch}{1.5}
\begin{tabular}{lcc}
\hline
Methods & F$_1$(\%) & mIoU(\%) \\
\hline
$p^* = \frac{a^T}{||a||} P^*$ & 72.9 & 79.1\\
$p^* = p + \alpha \frac{a^T}{||a||} P^*$ & 73.3 & 80.0\\
$p^* = \bar{a}^T P^*$ & 74.5 & 80.0 \\
$p^* = p + \alpha \bar{a}^T P^*$ & 74.8 & 81.1 \\
\hline
\end{tabular}
}
\end{minipage}
\end{figure}
\section{Conclusion}
This work leverages a tuning-free semantic segmentation approach based on SAM mask classification.
We utilize CLIP’s zero-shot classification capability to supplement sparse supervision or directly perform OVSS, enabling our method to adapt to various types of supervision with minimal adjustments.
The main challenge lies in CLIP's poor zero-shot classification performance, which leads to suboptimal pseudo-labels and OVSS results.
To address this, We propose DBA-CLIP, which fundamentally enhances CLIP's zero-shot classification by better aligning text and mask embeddings.
Additionally, we introduced GLCC to minimize pseudo-label noise interference by leveraging the intrinsic structure of mask embeddings.
Our method significantly outperforms the baseline model and advances the state-of-the-art across various benchmarks with different supervision types.

\clearpage

{\small
\bibliographystyle{ieee_fullname}
\bibliography{egbib}
}

\clearpage

\part*{Appendix}

\appendix

\setcounter{table}{0}   
\setcounter{figure}{0}
\setcounter{equation}{0}
\renewcommand{\thefigure}{S\arabic{figure}}
\renewcommand{\thetable}{S\arabic{table}}
\renewcommand{\thesection}{\Alph{section}}
\renewcommand{\theequation}{S\arabic{equation}}

\section{Approximation of DBA-CLIP}
\label{sec:proof-DAB-CLIP}
We approximate DBA-CLIP based on the following assumptions: 
1) Excluding the residual term $x$ does not significantly affect performance.
2) We disregard upsampling in $\mathcal{G}$ and assume that within a neighborhood of $x_f$, $\mathcal{G}$ behaves as a linear transformation $G \in R^{d \times d'}$. 
3) Single-head and multi-head self-attention have similar performance in our case.
Then we get:
\begin{align}
e &= \mathcal{N}( \bar{A}_{m}^T \mathcal{G}(x + x_f)), 
\label{eq:DBA-CLIP.Approx.bad}\\[1em]
&\approx \mathcal{N}( \bar{A}_{\hat{m}}^T x_f G^T),  
\notag \\[1em]
&= \mathcal{N}( \frac{1}{|\hat{m}|} \hat{m}^T A_{\hat{m}} x_f G^T), 
\notag \\[1em]
&\approx \mathcal{N}( \frac{1}{|m|} m^T \mathcal{G}(A_{\hat{m}} x_f)), 
\label{eq:DBA-CLIP.Approx.no_res} \\[1em]
&\approx \mathcal{N}( \frac{1}{|m|} m^T \mathcal{G}(x + A_{\hat{m}} x_f)), 
\label{eq:DBA-CLIP.Approx.good} \\[1em]
&\approx \mathcal{N}( \frac{1}{|m|} m^T \mathcal{G}(x + \mathrm{MHSA}(x))), 
\label{eq:DBA-CLIP.Approx.MHSA}
\end{align}
where
\begin{equation}
\label{eq:DBA-CLIP.Approx.bad:A_bar_m_hat_T}
\bar{A}_{\hat{m}}^T \in \mathbb{R}^{1 \times l} = \frac{1}{|\hat{m}|} \hat{m}^T A_{\hat{m}}
\end{equation}
denotes the average affinity at lower size, and $\mathrm{MHSA}(\cdot)$  represents the multi-head self-attention mechanism, in which each head operates identically to the single-head version of~\ref{eq:DBA-CLIP.Approx.good}. We ultimately adopted the multi-head version as it involves minimal modifications to the original CLIP.
\begin{table}[htpb]
\centering
\caption{
Performance comparison of the DBA-CLIP and its approximations on the PASCAL VOC 2012 \textit{trainaug} set.
}
\label{tab:DBA-CLIP.Approx}
\scalebox{1.0}{
\begin{tabular}{l|cc}
\hline
Methods & F$_1$(\%) & mIoU(\%) \\ 
\hline
\hline
Equation~\ref{eq:DBA-CLIP.Approx.bad} & 65.5 & 73.7 \\
Equation~\ref{eq:DBA-CLIP.Approx.no_res} & 66.6 & 74.2 \\
Equation~\ref{eq:DBA-CLIP.Approx.good} & 66.6 & 74.1 \\
Equation~\ref{eq:DBA-CLIP.Approx.MHSA} & 66.7 & 74.3 \\
\hline
\end{tabular}
}
\end{table}
We verify our assumptions in Table~\ref{tab:DBA-CLIP.Approx}. The results yielded by Equations~\ref{eq:DBA-CLIP.Approx.no_res},~\ref{eq:DBA-CLIP.Approx.good}, and~\ref{eq:DBA-CLIP.Approx.MHSA} are very close, indicating that the presence of residuals and the type of attention have little impact. 
Equation~\ref{eq:DBA-CLIP.Approx.bad} has a small mIoU gap compared to other methods and a larger gap in F$_1$, which is explained in Section~\ref{sec:DBA-CLIP}.

\section{Approximation of Inductive Label Propagation}
\label{sec:proof-GLCC}
Given the sparse affinity matrix $S$ among mask embeddings from the training set and the sparse affinity vector $a \in \mathbb{R}^{N}$ (defined in Equation~\ref{eq:GLCC.a}) of the testing embedding $e^v$ to the training embeddings $X$, the affinity matrix $\bar{S}$ for inductive label propagation can be approximated as:
\begin{align}
\bar{S} &\approx \begin{bmatrix}
S & a \\
a^T & 0 
\end{bmatrix} \\
\intertext{Here, We first assume that the $k$-nearest neighbors to the test embedding are symmetric, meaning that the test and training embeddings are always mutually $k$-nearest. Thus, the top-right partition of $\bar{S}$ is $a$. For the other $N-k$ training embeddings that do not rank among the $k$-nearest neighbors of the test embedding, their nearest neighbor search results remain consistent with those in the transductive scenario. Considering that the number of training samples $N$ is much larger than $k$, we can assume that the affinities among training samples remain unchanged, i.e., the top-left partition of $\bar{S}$ is still $S$. Then we have: }
\bar{S}+\bar{S}^T &= \begin{bmatrix}
S+S^T & 2 a \\
2 a^T & 0
\end{bmatrix} \\[1em]
\bar{D} & = \mathrm{diag}((\bar{S} + \bar{S}^T) \mathbf{1}) \approx \begin{bmatrix}
D & 0 \\
0 & 2 \sum_{i=1}^{N} a_i
\end{bmatrix} \\
\intertext{Here, we consider that $a$ is highly sparse, so the degree matrix of the training embedding part is still approximately equal to the degree matrix of $S + S^T$, that is, $D$. Next, we have:}
\label{eq:Why_D}
\bar{\mathcal{S}} &= \bar{D}^{-\frac{1}{2}}(\bar{S}+\bar{S}^T) \bar{D}^{-\frac{1}{2}} \\[1em]  \notag
&=\begin{bmatrix}
D & 0 \\
0 & 2 \sum_{i=1}^{N} a_i
\end{bmatrix}^{-\frac{1}{2}}\begin{bmatrix}
S+S^T & 2 a \\
2 a^T & 0
\end{bmatrix}\begin{bmatrix}
D & 0 \\
0 & 2 \sum_{i=1}^{N} a_i
\end{bmatrix}^{-\frac{1}{2}} \\[1em] \notag
&=\begin{bmatrix}
\mathcal{S} & \bar{a} \\
\bar{a}^T & 0
\end{bmatrix}, \ \text{where} \ \bar{a}_i=\frac{2 a_i}{\sqrt{2 D_{i i} \sum_{i=1}^{N} a_i}}=\frac{a_i}{\sqrt{\frac{1}{2} D_{i i} \sum_{i=1}^{N} a_i}} \\
\intertext{With Equation~\ref{eq:GLCC-bad}, we can get: }
& (\mathbf{I}-\alpha \bar{\mathcal{S}}) \begin{bmatrix}\bar{P}^* \\ p^*\end{bmatrix} = \begin{bmatrix}P^c \\ p\end{bmatrix} \\[1em] \notag
& \Rightarrow\begin{bmatrix}
\mathbf{I}-\alpha \mathcal{S} & -\alpha \bar{a} \\
-\alpha \bar{a}^T & 1
\end{bmatrix}\begin{bmatrix}\bar{P}^* \\ p^*\end{bmatrix} =\begin{bmatrix}P^c \\ p\end{bmatrix} \\[1em] \notag
& \Rightarrow \alpha p^* \bar{a} = (\mathbf{I}-\alpha \mathcal{S}) \bar{P}^* - P^c, \ \ p^*=p+\alpha \bar{a}^T \bar{P}^* \\
\intertext{We ignore the first term, which is difficult to solve. For the second term, we assume that:}
& \bar{P}^* \approx P^*,  \\
\intertext{which is quite reasonable since the inductive scenario only involves one more test embedding than the transductive scenario. There is no reason to believe that the inductive propagation result $\bar{P}^*$ on $N$ training embeddings would significantly differ from the transductive result $P^*$ due to this one extra test sample. Thus, we have: }
& p^*=p+\alpha \bar{a}^T P^*
\end{align}
Finally, we try to explain the meaning of $D_{ii}$ in the denominator of $\bar{a}_i$ from Equation~\ref{eq:Why_D} and why it is superior to plain L1 or no normalization used by~\cite{LP-LinearN,ZLaP}. In the embedding space, there are some dense subclusters (e.g., different parts of the same object), which can easily be included together during nearest neighbor searches. It is important to prevent these subclusters from dominating inductive label propagation results, as embeddings in them basically represent the same object. A key feature of embeddings from dense subclusters is their larger $D_{ii}$ due to high neighbor similarities. Therefore, we degree-normalize $a$ to reduce the impact of dense subclusters and increase neighbor diversity. Experiments in Table~\ref{tab:IL-Abl} show that this design outperforms L1 normalization.
\\ \\

\begin{figure}
    \centering
    \includegraphics[width=1.0\linewidth]{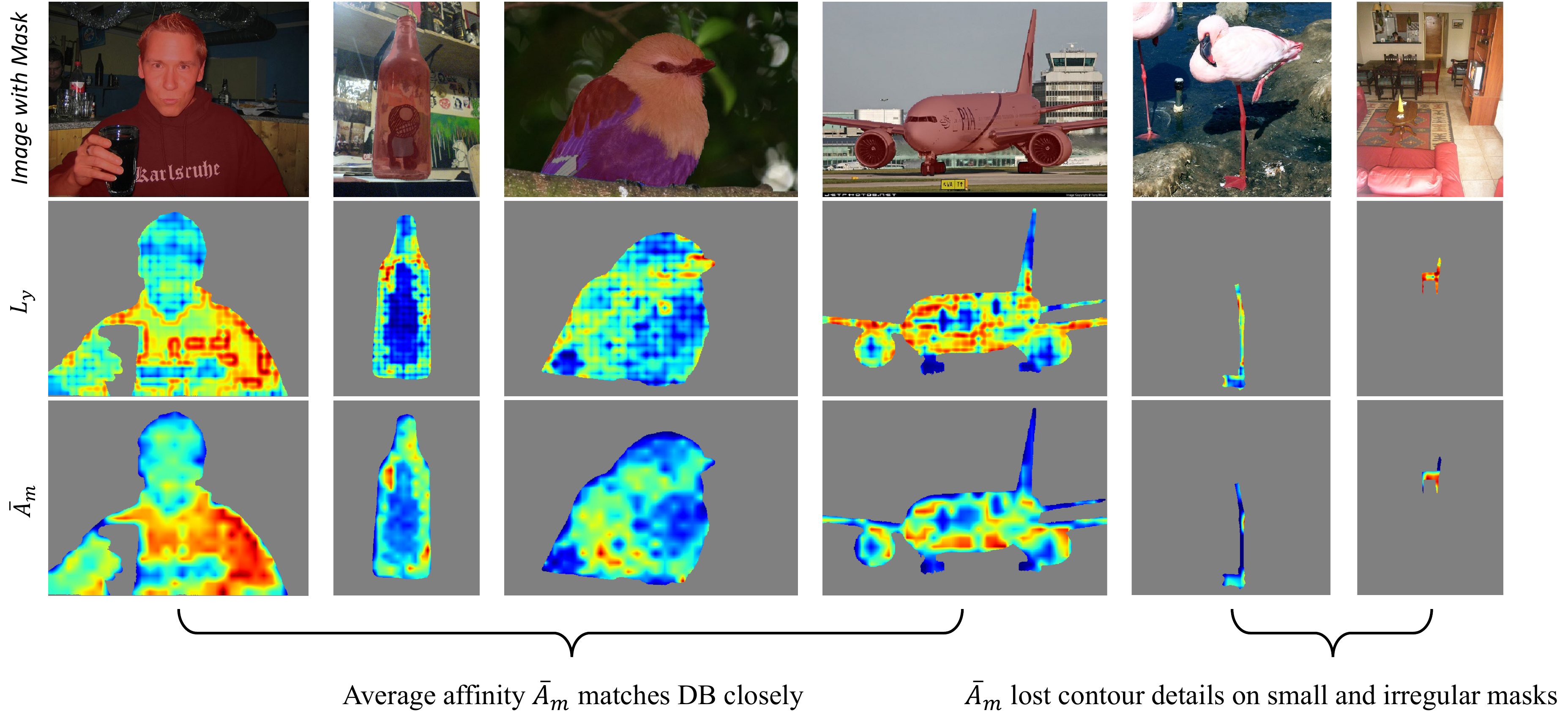}
    \caption{More examples of discrimination-bias alignment. 
    DB tends to overlook misleading areas like text and patterns, ambiguous regions such as faces, and indistinct zones like large solid colors. Instead, it focuses on the most discriminative regions.
    Average affinity aligns well with DB on large masks but becomes meaningless on small and irregular masks.}
    \label{fig:More-DBA}
\end{figure}

\begin{figure}
    \centering
    \includegraphics[width=1.0\linewidth]{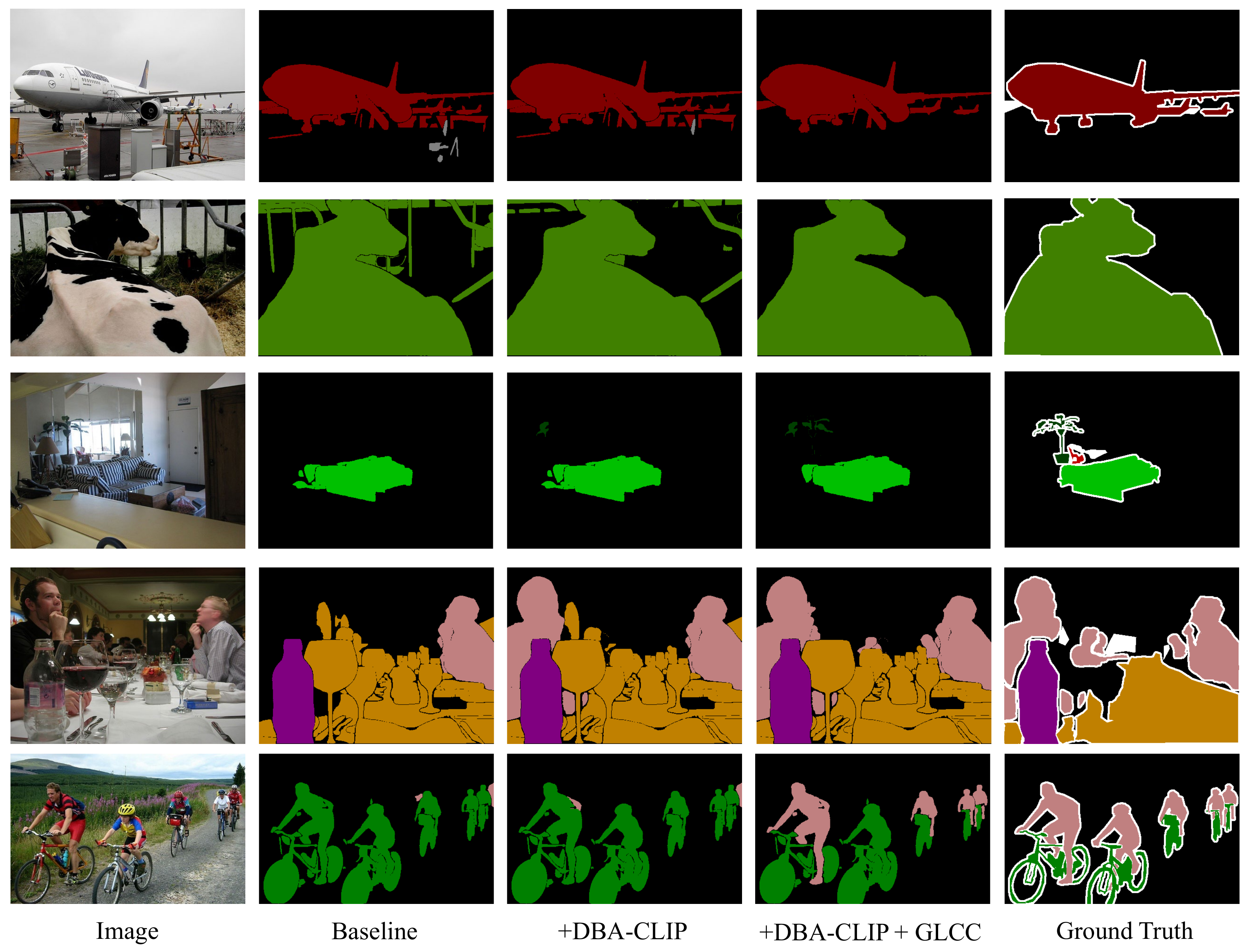}
    \caption{Illustration of segmentation results generated by the proposed method and its variants.}
    \label{fig:Seg}
\end{figure}

\begin{table}[t]
\centering
\caption{
Multi-stage weakly supervised semantic segmentation results on PASCAL VOC 2012 and MS COCO 2014.
The type of supervision is denoted in the "Sup." column, including full supervision ($\mathcal{F}$) and image-level labels ($\mathcal{I}$). The use of CLIP ($\mathcal{C}$), BLIP($\mathcal{B}$), Grounding DINO ($\mathcal{D}$)~\cite{GroundingDINO}, and SAM ($\mathcal{S}$) are also denoted. The \dag~indicates backbone pretrained on COCO~\cite{Lin2014COCO} for results on VOC, and \ddag~means ImageNet-21k~\cite{ImageNet-21k} pretrained backbone. Best results are marked in bold.
}
\label{tab:WSSS-SOTA-MS}
\begin{tabular}{lcclccc}
\toprule
\multirow{2}{*}{Methods} & \multirow{2}{*}{\textit{T.F.}} & \multirow{2}{*}{\textit{Sup.}} & \multirow{2}{*}{Backbone} & \multicolumn{2}{c}{VOC} & COCO \\
\cmidrule(lr){5-6} \cmidrule(lr){7-7}
& & & & \textit{val} & \textit{test} & \textit{val} \\
\midrule
\multicolumn{6}{l}{\textit{\textbf{Supervised Upperbounds}}} \\

DeepLab V3+{$_{\text{\color{gray}{CVPR18}}}$}~\cite{Chen2018deeplab} & \XSolidBrush & $\mathcal{F}$ & R101 & 79.5 & 79.4 & 60.4 \\

Mask2Former{$_{\text{\color{gray}{CVPR22}}}$}~\cite{mask2former} & \XSolidBrush & $\mathcal{F}$ & Swin-L\ddag & 86.0 & 86.1 & 66.7 \\

\midrule

\multicolumn{6}{l}{\textit{\textbf{Multi-stage Methods}}} \\

SIPE{$_{\text{\color{gray}{CVPR22}}}$}~\cite{Chen2022SPIE} & \XSolidBrush & $\mathcal{I}$ & R101\dag & 68.8 & 69.7 & 40.6 \\

CLIMS{$_{\text{\color{gray}{CVPR22}}}$}~\cite{Xie2022CLIMS} & \XSolidBrush & $\mathcal{I} + \mathcal{C}$ & R50\dag & 70.4 & 70.0 & - \\


Xu \etal{$_{\text{\color{gray}{CVPR23}}}$}~\cite{Xu2023} & \XSolidBrush &  $\mathcal{I} + \mathcal{C}$  & WR38 & 72.2 &72.2 & 45.9 \\

OCR+MCTformer{$_{\text{\color{gray}{CVPR23}}}$}~\cite{Cheng2023} & \XSolidBrush &  $\mathcal{I}$   & WR38 & 72.7 & 72.0 & 42.5 \\

BECO{$_{\text{\color{gray}{CVPR23}}}$}~\cite{Rong2023} & \XSolidBrush &  $\mathcal{I}$  & MiT-B2  & 73.7 &73.5 & 45.1\\

CLIP-ES{$_{\text{\color{gray}{CVPR23}}}$}~\cite{Lin2023CLIP-ES} & \XSolidBrush  & $\mathcal{I} + \mathcal{C}$  & R101\dag & 73.8 & 73.9 & 45.4\\

QA-CLIMS{$_{\text{\color{gray}{MM23}}}$}~\cite{QA-CLIPMS} & \XSolidBrush  & $\mathcal{I} + \mathcal{B}$  & WR38 & 75.6 & 75.5 & 43.2\\

Jiang \etal{$_{\text{\color{gray}{arXiv23}}}$}~\cite{Jiang2023SAMwsss} & \XSolidBrush   & $\mathcal{I} + \mathcal{S}$ & R101\dag & 71.1 & 72.2 & -\\

Sun \etal{$_{\text{\color{gray}{arXiv23}}}$}~\cite{Sun2023SAMwsss} & \XSolidBrush  & $\mathcal{I} + \mathcal{D} + \mathcal{S}$  & R101\dag & 77.2 & 77.1 & 55.6\\

WeakTr{$_{\text{\color{gray}{arXiv23}}}$}~\cite{WeakTr} & \XSolidBrush  &  $\mathcal{I}$  & ViT-S\ddag & 78.4 & 79.0 & 50.3\\

CLIP-ES + SEPL{$_{\text{\color{gray}{NIPS23}}}$}~\cite{SEPL} & \XSolidBrush  &  $\mathcal{I} + \mathcal{C} + \mathcal{S}$  & R101\dag & 73.1 & - & 47.9\\

FMA-WSSS{$_{\text{\color{gray}{WACV24}}}$}~\cite{FMA-WSSS} & \XSolidBrush  &  $\mathcal{I} + \mathcal{C} + \mathcal{S}$  & Swin-L\ddag & 82.6 & 81.6 & 55.4\\

CLIP-ES +CPAL{$_{\text{\color{gray}{CVPR24}}}$}~\cite{CPAL} & \XSolidBrush & $\mathcal{I} + \mathcal{C}$ & R101\dag & 74.5 & 74.7 & 46.8 \\

CoSA{$_{\text{\color{gray}{arXiv24}}}$}~\cite{CoSA} & \XSolidBrush  & $\mathcal{I}$ & Swin-B\ddag & 81.4 & 78.4 & 53.7 \\

\rowcolor{gray!40} \textbf{Ours} & \XSolidBrush  &  $\mathcal{I} + \mathcal{C} + \mathcal{S}$  & Swin-L\ddag & \textbf{84.0} & \textbf{84.2} & \textbf{57.3}\\ 

\bottomrule
\end{tabular}
\end{table}

\begin{table}[htpb]
\centering
\caption{
Performance comparison of the GLCC on the PASCAL VOC 2012 \textit{val} set under different hyper-parameter settings.
}
\label{tab:ak-Abl}
\resizebox{1.0\textwidth}{!}{
\begin{tabular}{ccccccccc}
\hline
\multirow{2}{*}{Metrics} & \multicolumn{4}{c}{$\alpha = 0.9$} & \multicolumn{4}{c}{$k = 50$} \\
\cmidrule(r){2-5} \cmidrule(r){6-9}
 & $k = 12$ & $k = 25$ & $k = 50$ & $k = 100$ & $\alpha = 0.5$ & $\alpha = 0.7$ & $\alpha = 0.9$ & $\alpha = 0.95$ \\
\hline
mIoU(\%) & 80.9 & \textbf{81.1} & \textbf{81.1} & 80.8 & 79.6 & 80.5 & \textbf{81.1} & 81.0 \\
F$_1$(\%) & 74.1 & 74.7 & \textbf{74.8} & 74.3 & 72.4 & 73.6 & \textbf{74.8} & 74.7 \\
\hline
\end{tabular}
}
\end{table}

\section{GLCC Efficiency}
\label{sec:GLCC-eff}
In Table~\ref{tab:efficiency}, we evaluate the construction and inference times of our newly introduced GLCC with two rounds of bootstrapping. The affinity matrix $\mathcal{S}$ needs to be computed only once and can then be reused for the transductive label propagation of each round. Unlike methods requiring fine-tuning, we need to infer each image in the training set just once, with a little extra time spent on constructing the GLCC. During testing, the overhead of GLCC is negligible compared to the inference of the foundation model. Note that we have not yet optimized the nearest neighbor search algorithm used. However, techniques such as quantization and partition~\cite{douze2024faiss,gpu-faiss} can significantly improve both construction and inference speeds when handling larger scales of training embeddings.

\begin{table}[htpb]
\centering
\caption{
Time for GLCC construction and inference on a single RTX 3090. "$N$" denotes the total number of mask embeddings extracted from the training set.
}
\label{tab:efficiency}
\resizebox{1.0\textwidth}{!}{
\begin{tabular}{lccccc}
\hline
\multirow{2}{*}{Dataset} & \multirow{2}{*}{$N$} & \multicolumn{3}{c}{Construction Time} & \multirow{2}{*}{Inference Time} \\
\cmidrule(r){3-5}
 & & Train Linear Probe & Compute $\mathcal{S}$ & Solve Eq.~\ref{eq:LP} & \\
\hline
VOC & 324,921 & 141s & 4s & <1s & 3.1 ms/img \\
COCO-Obj  & 4,660,424 & 1,075s & 732s & 22s & 13.6 ms/img \\
\hline
\end{tabular}
}
\end{table}

\section{Limitations}
\label{sec:limit}
Our method relies on the capabilities of the foundation model, which sets the upper bounds of our method. For example, in CityScapes~\cite{cordts2016cityscapes}, SAM~\cite{Kirillov2023SAM} often misses poles and fences and struggles to distinguish between roads and sidewalks. As a result, even if SAM masks are all correctly classified, it can only achieve an \textasciitilde80\% mIoU. Therefore, fine-tuning the base model is still recommended for datasets with uncommon annotation preferences, highly challenging and diverse large-scale data, or scenes rarely seen in web images (such as industrial or medical data). However, this does not prevent combining our framework with fine-tuning. For instance, we can fine-tune CLIP~\cite{Radford2021CLIP} with image-level labels on new data and then use our framework to obtain semantic segmentation results.

Another concern is that the high inference cost of foundational models may limit the deployability. 
On one hand, we anticipate advancements in more efficient foundational models~\cite{MobileCLIP,MobileSAM}.
On the other hand, for WSSS or SSS, our method can infer on the training set to generate pseudo-masks and then train a standard semantic segmentation network on them. This approach should offer greater deployment flexibility, although it sacrifices the advantage of being tuning-free.

\clearpage
\newpage

\end{document}